\documentclass[10pt,twocolumn,letterpaper]{article}

\usepackage{iccv}
\usepackage{times}
\usepackage{epsfig}
\usepackage{graphicx}

\usepackage{helvet,courier}
\usepackage[tight,footnotesize]{subfigure}
\usepackage{amsmath,amssymb,amsthm,bm}
\usepackage{multirow,makecell,footmisc,url}
\usepackage[ruled]{algorithm2e}
\usepackage{color}

\newcommand{\redbold}[1]{\textcolor{red}{\textbf{#1}}}
\newcommand{\bluebold}[1]{\textcolor{blue}{\textbf{#1}}}

\usepackage[pagebackref=true,breaklinks=true,letterpaper=true,colorlinks,bookmarks=false]{hyperref}

\iccvfinalcopy % *** Uncomment this line for the final submission

\def\iccvPaperID{}

\ificcvfinal\fi
\begin{document}

%%%%%%%%% TITLE
\title{Spatial-Temporal Relation Networks for Multi-Object Tracking}

\author{    
Jiarui Xu$^{13}$\thanks{This work is done when Jiarui Xu and Yue Cao are interns at Microsoft Research Asia.}, Yue Cao$^{23}$, Zheng Zhang$^3$, Han Hu$^3$\\
    $^1$Hong Kong University of Science and Technology\\
    $^2$School of Software, Tsinghua University\\
    $^3$Microsoft Research Asia\\
    {\tt \small jxuat@ust.hk, caoyue10@gmail.com, \{zhez,hanhu\}@microsoft.com}
}

\maketitle

\begin{abstract}
Recent progress in multiple object tracking (MOT) has shown that a robust similarity score is key to the success of trackers. A good similarity score is expected to reflect multiple cues, e.g. appearance, location, and topology, over a long period of time. However, these cues are heterogeneous, making them hard to be combined in a unified network. As a result, existing methods usually encode them in separate networks or require a complex training approach. In this paper, we present a unified framework for similarity measurement which could simultaneously encode various cues and perform reasoning across both spatial and temporal domains. 
We also study the feature representation of a tracklet-object pair in depth, showing a proper design of the pair features can well empower the trackers. The resulting approach is named spatial-temporal relation networks (STRN). It runs in a feed-forward way and can be trained in an end-to-end manner. The state-of-the-art accuracy was achieved on all of the MOT15$\sim$17 benchmarks using public detection and online settings.
\end{abstract}

\vspace{-1em}
\section{Introduction}

\begin{figure}[tbp]
  \centering
  \includegraphics[width=0.43\textwidth]{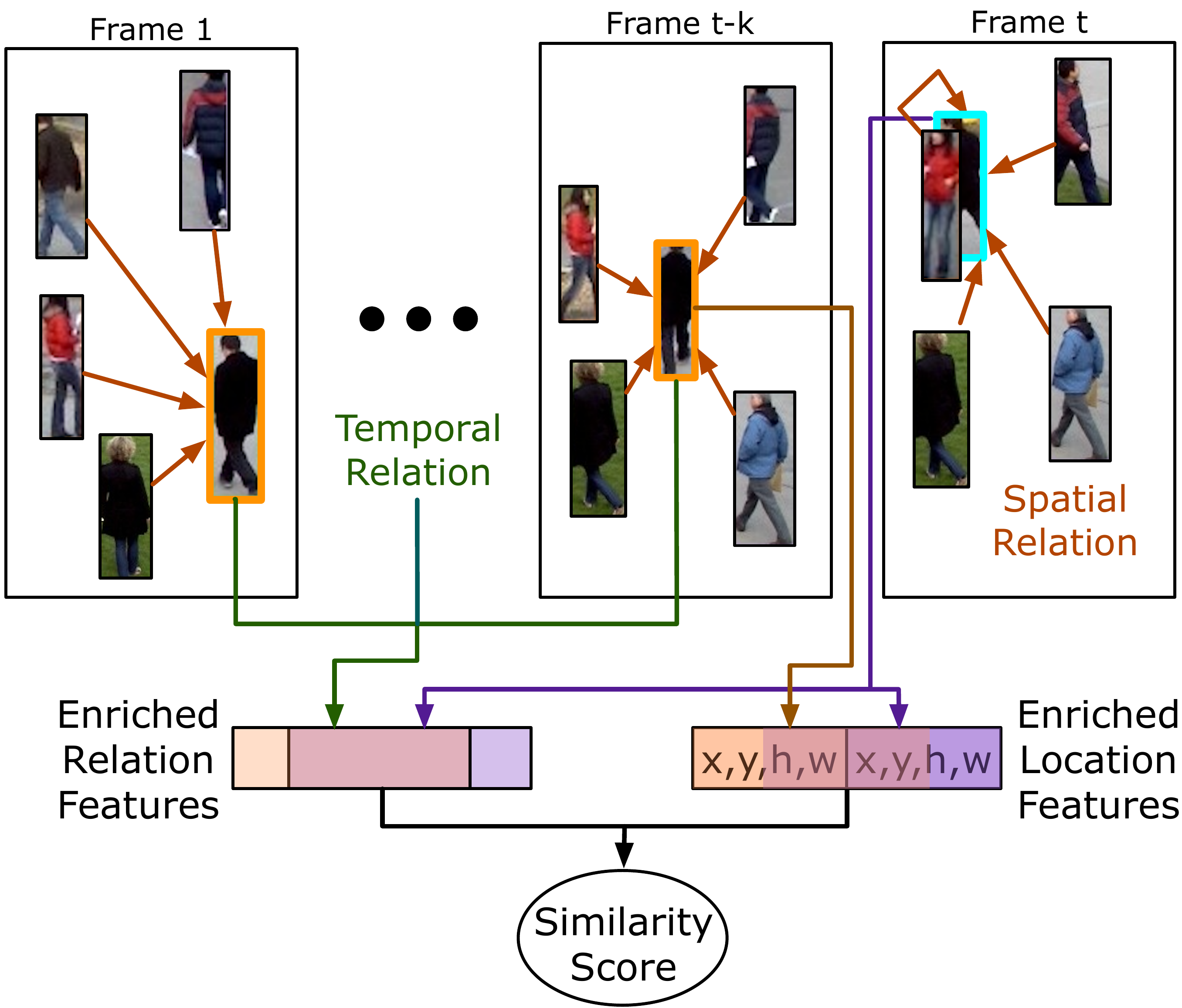}
  \caption{The proposed spatial-temporal relation networks (STRN) to compute similarity scores between tracklets and objects. The networks can combine various cues such as appearance, location, and topology, and aggregation information over time. The orange boxes and the blue box indicate the same person in different frames.}
  \label{fig:teaser}
  \vspace{-2pt}
\end{figure}

Multiple object tracking (MOT) aims at locating objects and maintaining their identities across video frames. It has attracted a lot of attention because of its broad applications such as surveillance, sports game analysis, and autonomous driving. Most recent approaches follow the popular ``tracking-by-detection'' paradigm \cite{chu2017online,hong2016online,leal2016learning,maksai2017non,milan2017online,matilla2016strongweak,xiang2015learning}, where objects are firstly localized in each frame and then associated across frames. Such a decoupled pipeline reduces the overall complexity and shifts the major attention of MOT to a more unitary problem: object association. This paradigm also benefits from the rapid progress in the field of object detection \cite{ross2010dpm,ren2015faster,yang2016sdp,dai2017deformable} and has led several popular benchmarks for years, i.e. MOT15$\sim$17 \cite{MOTChallenge2015,MOT16}. 

In general, the performance of object association highly depends on a robust similarity score. The similarities in the most existing approaches are only based on the appearance features extracted from the cropped object patches~\cite{leal2017tracking}. The performance by such similarities is limited due to the following reasons: Firstly, the objects are often from the same category in tracking scenario, e.g. \emph{person} in MOT15$\sim$17 benchmark, with appearance hard to be distinguished. Secondly, objects across frames also suffer from frequent occlusions and quality/pose variations, which further increases the difficulty in building a robust similarity score.

The pioneering works of exploring varying cues to build the similarity score have been proven to be effective\cite{sadeghian2017tracking,chu2017online,zhu2018online,xiang2015learning}. Convolutional neural networks have been well studied and employed to encode appearance cue \cite{wang16joint,zhu2018online}, and the hand-crafted location cues are integrated with appearance cue in recent works~\cite{sadeghian2017tracking,chu2017online,zhu2018online}. The topological structure~\cite{sadeghian2017tracking} between bounding boxes is crucial for judging whether a pair of bounding boxes in different frames indicate the same object, especially for occlusion. As shown in Figure~\ref{fig:teaser}, the orange bounding boxes in frame $1$ and frame $t-k$ and blue bounding box in frame $t$ indicate the same person. Although the person in frame $t$ has obscured by another person, and its appearance has a great difference compared with previous frames, the topological information keeps consistent and makes the obscured person identifiable. Besides, aggregation information across frames is also verified to be beneficial for measuring similarity~\cite{sadeghian2017tracking, kim2018multi,milan2017online}. 

However, because of the \emph{heterogeneous} representation of different cues and resulting in the difficulties of dealing with all the cues into one unified framework, these works are usually based on cue-specific mechanisms~\cite{sadeghian2017tracking,kim2018multi,milan2017online, kim2018multi} and required sophisticated learning approaches~\cite{sadeghian2017tracking}. For example, \cite{sadeghian2017tracking} uses an occupancy map to model topological information and ~\cite{kim2018multi} uses a specialized gating mechanism in RNN to aggregate information over time.

Our work is motivated by the success of relation networks in natural language problems~\cite{vaswani2017attention} and vision problems~\cite{hu2017relation,wang2017non,battaglia2016interaction,santoro2017simple}. In the relation networks, each element aggregates features from other elements through a content-aware aggregation weight, which can be automatically learned according to the task goal without explicit supervision. Since there is not an excessive assumption about the data forms, the relation networks are widely used in modeling dependencies between distant, non-grid or differently distributed data, such as word-word relation~\cite{vaswani2017attention}, pixel-pixel relation~\cite{wang2017non} and object-object relation~\cite{hu2017relation, battaglia2016interaction, santoro2017simple}. These data forms are hard to be modeled by regular convolution or sequential networks.

In this paper, we present a unified framework for similarity measurement by integrating multiple cues in an end-to-end manner through extending the object-object relation network~\cite{hu2017relation} from the spatial domain to the spatial-temporal domain. With the extension of relation networks, we elegantly encode the appearance and topology cues for both objects and tracklets. It is able to accommodate location-based cues as well. 

The whole module is illustrated in Figure~\ref{fig:teaser}. Our goal is to compute the similarity between objects in the current frame and referenced tracklets on previous frames. The spatial-temporal relation networks are firstly applied in each frame to strengthen the appearance representation of an object in the spatial domain. Then, the strengthened features on its referenced tracklet are aggregated across time via applying our relation networks in the temporal domain. Finally, the aggregated features on the tracklet and the strengthened features of the object are concatenated to enrich the representation of the tracklet-object pair and produce a similarity score accordingly. We also show that the proper design of feature representation for the tracklet-object pair is crucial for the quality of similarity measure. The resulting approach is named spatial-temporal relation networks (STRN), which is fully feed-forward, can be trained in an end-to-end manner and achieves state-of-the-art performance over all online methods on MOT15$\sim$17 benchmarks.

\section{Related Work}
\textbf{Tracking-by-Detection Paradigm  } 
Recent multiple object tracking (MOT) methods are mostly based on the tracking-by-detection paradigm, with the major focus on the object association problem. According to what kind of information is used to establishing the association between objects in different frames, the existing methods can be categorized into online methods~\cite{bae2014robust,chu2017online,hong2016online,leal2016learning,maksai2017non,milan2017online,oh2009mcmc,matilla2016strongweak,shu2012part,xiang2015learning,yang2012multi,chen2017online}, and offline methods~\cite{brendel2011multiobject,dehghan2015gmmcp,milan2014continuous,pirsiavash2011globally,Pirsiavash2011CVPR,ZamirECCV12,son2017multi,tang2015subgraph,tang2016multi,zhang2008global}. The former methods are restricted to utilize past frames only in the association part, which is consistent with real-time applications. The latter methods can incorporate both past and future frames to perform more accurate association.

Our method also follows the tracking-by-detection paradigm and mainly focus on improving the measurement of object similarities. For better illustration and comparison with other methods, we only instantiate the online settings in this paper, but the proposed method is also applicable to both offline and online association. 

\textbf{Similarity Computation  }
The major cues to compute similarities include appearance, location and topology.

The evolution of appearance feature extractor is from hand-craft \cite{bae2014robust,oh2009mcmc,shu2012part,yang2012multi} to deep networks \cite{zhu2018online,sadeghian2017tracking,chu2017online,Ren18CDRL,kim2018multi}. In this paper, we also utilize deep networks as our base appearance feature extractor. One crucial difference between the previous approaches lies in the way to build similarity from appearances. We utilize a hybrid of feature concatenation, cosine distance, location/motion priors to compute the final similarities.

The utilization of location/motion features is common as well. Most existing methods assume a prior motion model, such as slow velocity \cite{Bochinski17highspeed} and linear/non-linear motion model \cite{zhu2018online}.
For example, the IoU trackers \cite{Bochinski17highspeed} rely on the assumption that objects in consecutive frames are expected to have high overlap, which is often not hold by fast moving objects. Other hard motion models also face the similar problem resulting in limited application. 
In this paper, instead of using hard location/motion priors, we integrate both unary location and motion information into the feature and learn the soft location/motion representation from data. Empirical studies on several benchmarks have proved the effectiveness of the learnable location representation. 

The topological information is also crucial for measuring similarity~\cite{sadeghian2017tracking}. However, leveraging such non-grid topology of multiple objects is challenging. Only a few works successfully encode the topological information, e.g. the occupancy grid in~\cite{sadeghian2017tracking}. 
However, this occupancy grid only counts the distribution of objects, without differentiating individual objects. In this paper, we utilize relation networks to encode the topological information for making the individual object differentiable and identifiable.

Most existing methods utilize one or two cues for similarity computation, while only a few works trying to jointly learn all of them simultaneously~\cite{sadeghian2017tracking}. 
Aggregating information across time~\cite{chu2017online,sadeghian2017tracking,kim2018multi,zhu2018online} is also rare. 
In addition, in order to learn the representations of different cues, these works usually adopt separate networks and sophisticated training strategy, e.g. a four-stage training is required by~\cite{sadeghian2017tracking}. 

In this paper, we combine all of the mentioned cues across time for similarity measurement by using a unified framework, which is fully feed-forward and it can be trained in end-to-end. In addition to the cues representing individual objects, we also study the representation for a tracklet-object pair in depth. We find that proper design of the pair representation is crucial for the quality of measuring similarity.

\textbf{Relation Networks  } Recently, relation networks have been successfully applied in the fields of NLP, vision and physical system modeling~\cite{hu2017relation,vaswani2017attention,wang2017non,battaglia2016interaction,santoro2017simple}, in order to capture long-term, non-grid and heterogeneous dependencies between elements.

Our approach is motivated by these works by extending the relation networks to multi-object tracking. In order to model the topological information of objects in the spatial domain and perform information aggregation over the temporal domain, we propose a spatial-temporal relation network. 
Although some recent works~\cite{chu2017online,zhu2018online} attempt to incorporate the attention mechanism into the multi-object tracking problem, they mainly aim at recovering salient foreground areas within a bounding box, thus alleviating the occlusion problem and ignoring the topology between objects.

\section{Method}

The goal of multi-object tracking (MOT) is to predict trajectories of multiple objects over time, denoted as ${\bf{T}} = {\left\{ {{{\bf{T}}_i}} \right\}_{i = 1}^N}$.
The trajectory of the $i^\text{th}$ object can be represented by a series of bounding boxes, denoted by ${{\bf{T}}_i} = {\left\{ {{\bf{b}}_i^t} \right\}_{t = 1}^T},{\bf{b}}_i^t = \left[ {x_i^t,y_i^t,w_i^t,h_i^t} \right]$. $x_i^t$ and $y_i^t$ denote the center location of the target $i$ at frame $t$. $w_i^t$ and $h_i^t$ denote the width and height of the target object $i$, respectively.

\begin{figure}[tbp]
  \centering
  \includegraphics[width=0.48\textwidth]{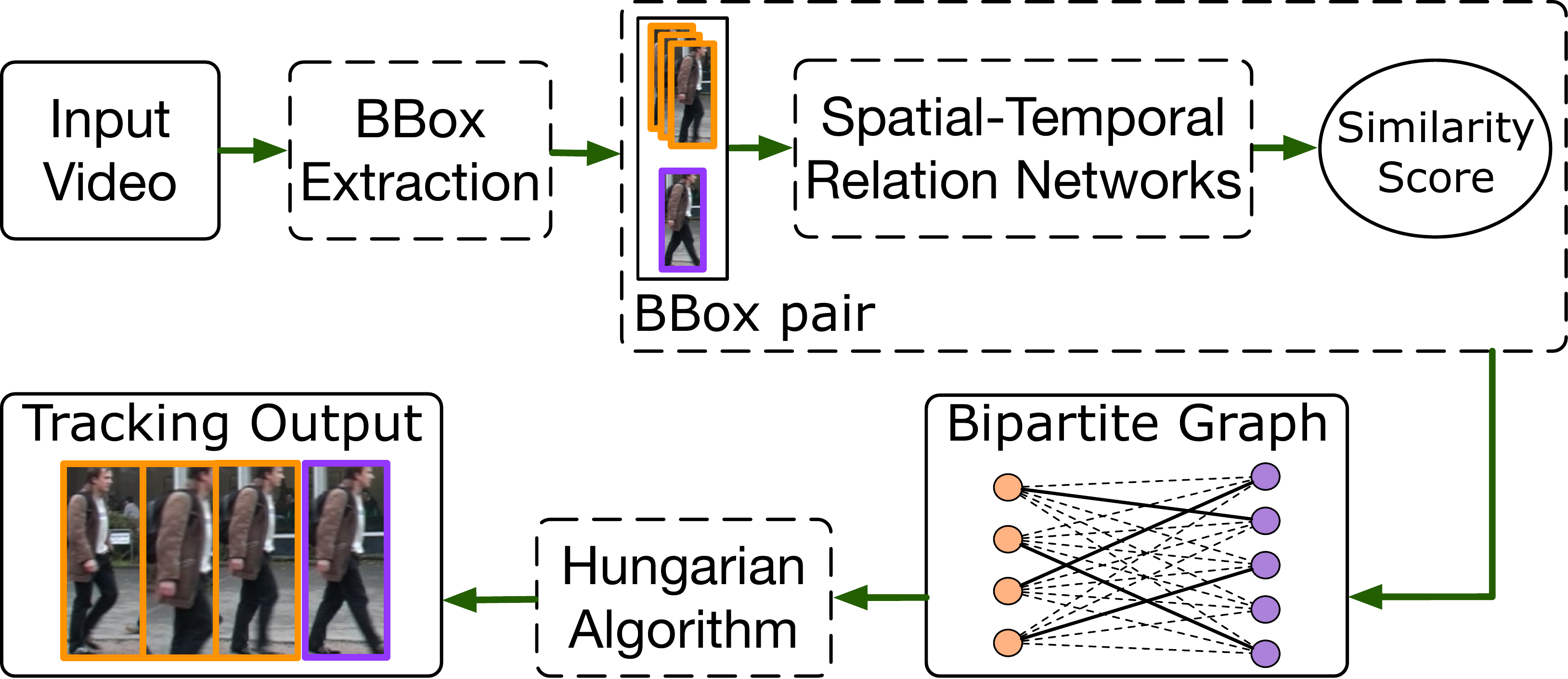}
  \caption{The online tracking-by-detection pipeline for multi-object tracking.}
  \label{fig:pipeline}
\end{figure}

Our method follows the online tracking-by-detection paradigm~\cite{xiang2015learning}, which first detects multiple objects in each frame and then associates their identities across frames.
The pipeline is illustrated in Figure~\ref{fig:pipeline}.
Given a new frame with the detected bounding boxes, the tracker computes similarity scores between the already obtained tracklets and the newly detected objects, resulting in a bipartite graph. Then the Hungarian algorithm~\cite{munkres1957algorithms} is adopted to get the optimal assignments. When running the assignment process frame-by-frame, object trajectories are yielded.

This paper is mainly devoted to building the robust similarity scores between tracklets extracted in previous frames and objects on the current frame, which proves crucial for multi-object tracking~\cite{leal2017tracking}. Denote the $i^\text{th}$ tracklet before frame $t-1$ as ${{\bf{T}}_i^{t-1}} = \left\{ {{\bf{b}}_i^1,{\bf{b}}_i^2,...,{\bf{b}}_i^{t - 1}} \right\}$ and the extracted objects at current frame $t$ as ${{\bf{D}}_t} = \left\{ {{\bf{b}}_j^t} \right\}_{j = 1}^{{N_t}}$. Each pair $\left({{\bf{T}}_i^{t-1}}, {\bf{b}}_j^t\right)$ is assigned a similarity score $s_{ij}^t$.

\begin{figure*}[tbp]
  \centering
  \includegraphics[width=0.85\textwidth]{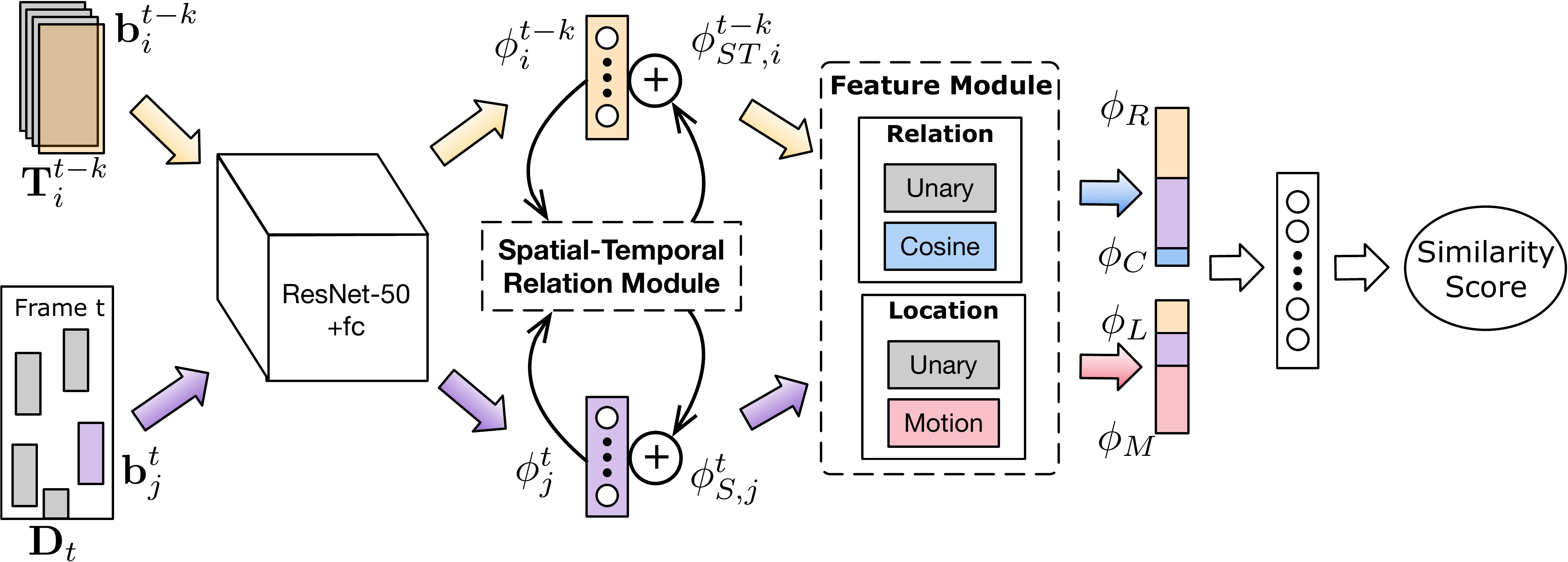}
  \caption{The architecture of Spatial-Temporal Relation Networks (STRN) to compute similarity scores between tracklets and objects.}
  \label{fig:structure}
  \vspace{-5pt}
\end{figure*}

As mentioned before, the appearance, location, topology cues, and aggregating information over time are all useful in computing the similarity scores. In this paper, we present a novel method based on spatial-temporal relation networks to simultaneously encode all of the mentioned cues and perform reasoning across time. Figure \ref{fig:structure} summarizes the entire process of similarity computation. Firstly, basic appearance features are extracted by a deep neural network, i.e. ResNet-50 in this paper, for both objects on current frame and objects on already obtained tracklets in previous frames, denoted as $\phi_i^t$ (object $i$ on frame $t$). Then the appearance features of objects across space and time are reasoned through a spatial-temporal relation module (STRM), resulting in spatial strengthened representation and temporal strengthened representation, denoted as $\phi_{\text{S}, i}^t$ and $\phi_{\text{ST}, i}^t$, respectively. Through these two strengthened features, we further develop the two types of relationship features $\phi_R$ and $\phi_C$ by concatenating them together and calculating the cosine distance between them, respectively. Finally, we combine the relation features with the unary \emph{location} feature $\phi_L$ and \emph{motion} feature $\phi_M$ together as the representation of a tracklet-object pair $\left({{\bf{T}}_i^{t-k}}, {\bf{b}}_j^t\right)$. Accordingly, the similarity is obtained by a two-layer network with a sigmoid function. 

In the following subsections, we will present each mentioned feature in detail. We firstly introduce the spatial-temporal relation module (STRM), which acts as a central role in combining various cues and performing reasoning across time. Then we present the design of the feature presentation for a tracklet-object pair, which proves crucial for the quality of similarity measure.

\subsection{Spatial-Temporal Relation Module}\label{sec:relation}

We firstly review the basic \emph{object relation module}, which is introduced in~\cite{hu2017relation} to encode context information for object detection.
\vspace{-0.5em}
\paragraph{Object relation module (ORM)} The basic object relation module~\cite{hu2017relation} aims at strengthening an input appearance feature by aggregating information from other objects within a static image(a static image is a single frame in video). We denote object by ${o_i} = \left( {{\phi _i},{{\mathbf{b}}_i}} \right)$, with $\phi _i$ the input appearance feature and $\mathbf{b}_i=(x_i,y_i,w_i,h_i)$ the object location. The object relation module computes a refined feature of object $o_i$ by aggregating information from an object set $\mathcal{O} = \left\{ {{o_j}} \right\}_{j = 1}^N = \left\{ {\left( {{\phi _j},{{\bf{b}}_j}} \right)} \right\}_{j = 1}^N$:
\begin{small}
\begin{equation} \label{eqn:relation-func}
{\phi_i'} = {\phi_i} + \sum\limits_j {{\omega _{ij}} \cdot \left( {{W_V} \cdot {\phi _j}} \right)},
\end{equation}
\end{small}
where $\omega_{ij}$ is the attention weight contributed from object $o_j$ to $o_i$; $W_V$ is a transformation matrix of the input features.

Attention weight $\omega_{ij}$ is computed considering both the projected appearance similarity $\omega_{ij}^A$ and a geometric modulation term $\omega_{ij}^G$ as
\begin{small}
\begin{equation}\label{eqn:weight}
{\omega _{ij}} = \frac{{\omega _{ij}^G \cdot \exp \left( {\omega _{ij}^A} \right)}}{{\sum\limits_{k = 1}^N {\omega _{ik}^G \cdot \exp \left( {\omega _{ik}^A} \right)} }}.
\end{equation}
\end{small}
$\omega_{ij}^A$ is denoted as the scaled dot product of projected appearance features ($W_Q$, $W_K$ are the projection matrices and $d$ is the dimension of projected feature)~\cite{vaswani2017attention}, formulated as
\begin{small}
\begin{equation}\label{eqn:appearance}
	\omega _{ij}^A = \frac{{\left\langle {{W_Q}{\phi _i},{W_K}{\phi _j}} \right\rangle }}{{\sqrt {d} }}.
\end{equation}
\end{small}
$\omega_{ij}^G$ is obtained by applying a small network on the relative location $\log \left( {\frac{{\left| {{x_i} - {x_j}} \right|}}{{{w_j}}},\frac{{\left| {{y_i} - {y_j}} \right|}}{{{h_j}}},\frac{{{w_i}}}{{{w_j}}},\frac{{{h_i}}}{{{h_j}}}} \right)$.
The original object relation module in~\cite{hu2017relation} only performs reasoning within the spatial domain. In order to better leverage the advantage of object relation module in multi-object tracking, we extend this module to the temporal domain in this paper.  

\vspace{-1.0em}
\paragraph{Extension to the spatial-temporal domain}
The object relation module can be extended to the spatial-temporal domain in a straight-forward way by enriching the object set $\mathcal{O}$ by all objects from previous frames. Such solution is obviously sub-optimal: firstly, the complexity is significantly increased due to more objects involved in reasoning; secondly, the spatial and temporal relations are tackled with no differentiation. In fact, spatial and temporal relations are generally expected to contribute differently to the encoding of cues. The spatial relation could draw on strengths in modeling \emph{topology} between objects. The temporal relation is fit for aggregating information from multiple frames, which could potentially avoid the degradation problem caused by accidental low-quality bounding boxes.

Regarding the different effects of spatial and temporal relations, we present a separable spatial-temporal relation module, as illustrated in Figure~\ref{fig:teaser}. It firstly performs relation reasoning in the spatial domain on each frame. The spatial reasoning process strengthens input appearance features with automatically learned topology information. Then the strengthened features on multiple frames are aggregated through a temporal relation reasoning process.

The spatial relation reasoning process strictly follows equation 1 to encode topological clues, and the output characteristic of the process is expressed as p, whose encoded topological structure has been proved to be effective in the field of object detection.

\begin{figure*}[!htb]
	\centering
		\includegraphics[width=0.9\textwidth]{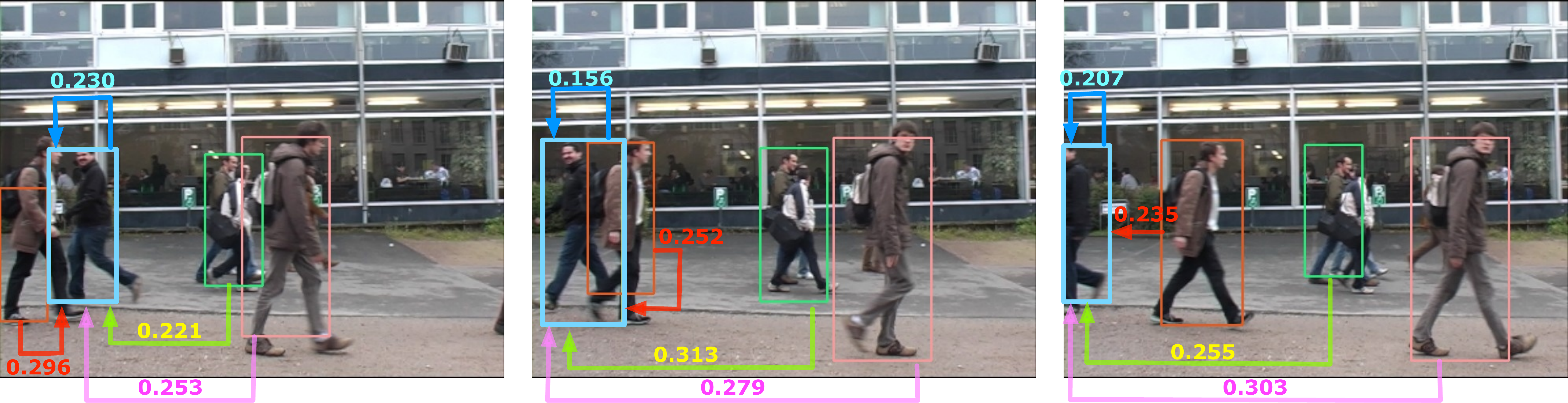}
	\caption{
		Learnt spatial attention weights across frames.}
	\label{fig:spatial_att_vis}
	\vspace{-1.0em}
\end{figure*}

The two types of relations follow different formulations. The spatial relation reasoning process strictly follows Eqn.~(\ref{eqn:relation-func}) to encode the \emph{topology} cue and the resulting output feature of this process is denoted as $\phi_{S,i}$, which has been proved to be effective in encoding the \emph{topology} information to improve object detection~\cite{hu2017relation}. Figure~\ref{fig:spatial_att_vis} illustrates the learnt spatial attention weights across frames. In general, the attention weights are stable on different frames, suggesting certainly captured topology representation. It should be noticed that the attention weight of an object itself is not necessarily higher than others, since $W_{Q}$ and $W_{K}$ in Eqn.~(\ref{eqn:relation-func}) are different projections. This is also the case for geometric weights.

\begin{figure}[!htb]
	\centering
		\includegraphics[width=0.45\textwidth]{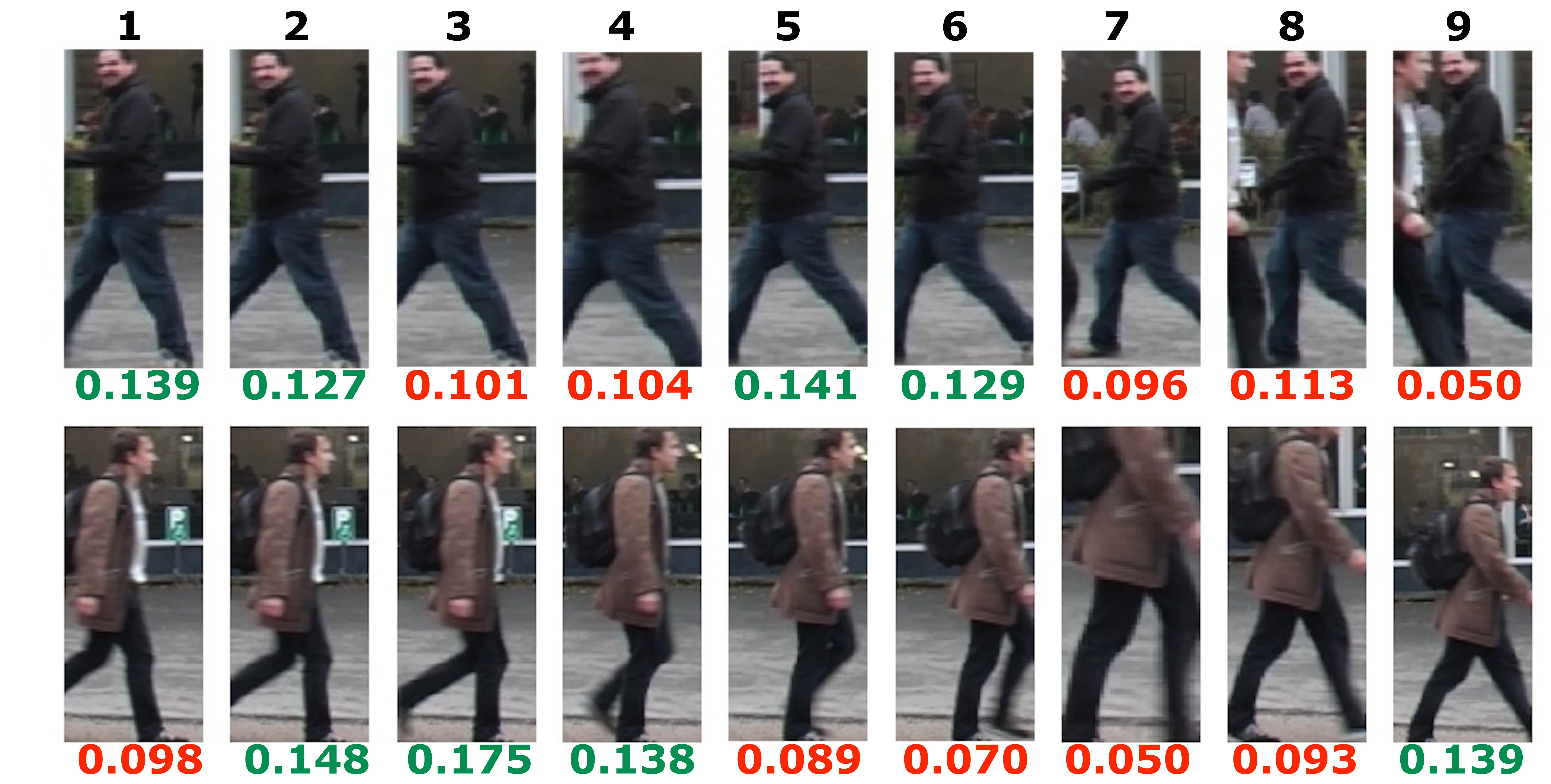}
	\caption{
		Learnt temporal attention weights.}
	\label{fig:temporal_att_vis}
	\vspace{-1.0em}
\end{figure}

The temporal relation reasoning process is conducted right after spatial relation reasoning\footnote{Note the temporal relation reasoning is only performed for tracklets. The encoding of objects on current frame only includes spatial reasoning.}. Instead of strengthening particular object features on each frame as in spatial relation modeling, we compute a representation of the whole tracklet by aggregating features from multiple frames. Due to the limiting of memory, the aggregation is only performed on latest $\tau_{1}$ frames($\tau_{1}=10$ in default):
\begin{small}
\begin{equation} \label{eqn:temporal_relation}
{\phi_{ST, i}^t} = \sum\limits_{k=0}^{\tau_{1}-1} {{\omega_{i}^{t-k}} \cdot {\phi _{S, i}^{t-k}}}.
\end{equation}
The attention weight is defined on the individual input feature as
\begin{equation} \label{eqn:temporal_attention_weight}
\omega_i^t = \frac {\exp(\left\langle \mathbf{w}_T, \phi_{S,i}^t\right\rangle)} {\sum\limits_k \exp ( \left\langle \mathbf{w}_T, \phi_{S,i}^k\right\rangle)}.
\end{equation}
\end{small}
Eqn.~(\ref{eqn:temporal_relation}) is essentially a weighted average of object features from recent frames. The learnt temporal attention weights is illustrates in Figure~\ref{fig:temporal_att_vis}. The blurring, wrongly cropped or partly occluded detections are assigned with low attention weights, indicating feature qualities are automatically learnt, and the representation of a tracklet will be less affected by these low quality detections.

\vspace{-0.5em}
\subsection{Design of Feature Representation} \label{sec:feature}
The performance of a practical vision system highly relies on the proper design of feature representation. The previous subsection mainly discuss encoding the cues of an individual object or tracklet. This section studies the representation of the tracklet-object pair in depth. Specifically, the \emph{relation features} produced by the spatial-temporal relation module and the \emph{location features} which represent the geometric properties of bounding boxes are combined together to form the representation of a tracklet-object pair.

In general, the features of tracklets and objects suffer certain incompatibility since the feature representation of tracklets involve the temporal reasoning process while the features of objects not. To tackle such incompatibility issue, we follow recent practice~\cite{sadeghian2017tracking,zhu2018online} to computing the similarity. The concatenated features of tracklets and objects are fed into a two-layer network followed by a sigmoid function to producing similarity score, as
\begin{small}
\begin{equation}
\label{eqn:relation_concatenation}
s_{ij}^t = \text{sigmoid} \left(W_{s2} \cdot \text{ReLU}(W_{s1} \cdot [\phi_{R};\phi_{C};\phi_{L}; \phi_{M}])\right),
\end{equation}
\end{small}
where $\phi_{R}$ (in Eqn. \ref{eqn:relation-phi}) denotes relation features, $\phi_{C}$ (in Eqn. \ref{eqn:relation_cosine}) denotes the cosine similarity, $\phi_{L}$ (in Eqn. \ref{eqn:absolution_location}) denotes location features and $\phi_{M}$ (in Eqn. \ref{eqn:motion_vector}) denotes motion features.

\subsubsection{Relation Features}
The spatial relation module couple the appearance cue and topology cue of an object. The temporal relation module aggregating information across frames. 

Since an object corresponded tracklet may exceed the image boundary, or be lost tracked due to the imperfection of the system, the tracklet does not necessarily appear at last frame. We need to enlarge the candidate tracklets from the last frame to multiple frames. Because of the memory limiting, only recent $\tau_2$ frames are involved($\tau_2$=10 in default).

We directly perform a linear transform on input relation features, which are regarded as the base feature type.

\begin{small}
\begin{equation}\label{eqn:relation-phi}
    \phi_{R} = W_R \cdot \left[\phi_{ST,i}^{t-k};\phi_{S,j}^t\right], 1\leq k \leq \tau_2
\end{equation}
\end{small}
where $W_R$ is a linear transform for feature fusion.

Directly using the concatenated relation features enables computing similarity of different modes. However, the freedom in representation is double-edged that it also increases the difficulty in learning compact individual features.

To address this issue, we propose to explicitly compute the cosine similarity between two relation features:
\begin{small}
\begin{equation}
\label{eqn:relation_cosine}
\phi_{C} = \text{cos} \left({{W_C} \cdot \phi _{ST,i}^{t - k},{W_C} \cdot \phi _{S,j}^t} \right), 1\leq k \leq \tau_2
\end{equation}
\end{small}
where $W_C$ is a linear layer to project the original relation features into a low-dimensional representation, e.g. 128-d. 

\begin{figure}[!tb]
    \centering
        \includegraphics[width=0.3\textwidth]{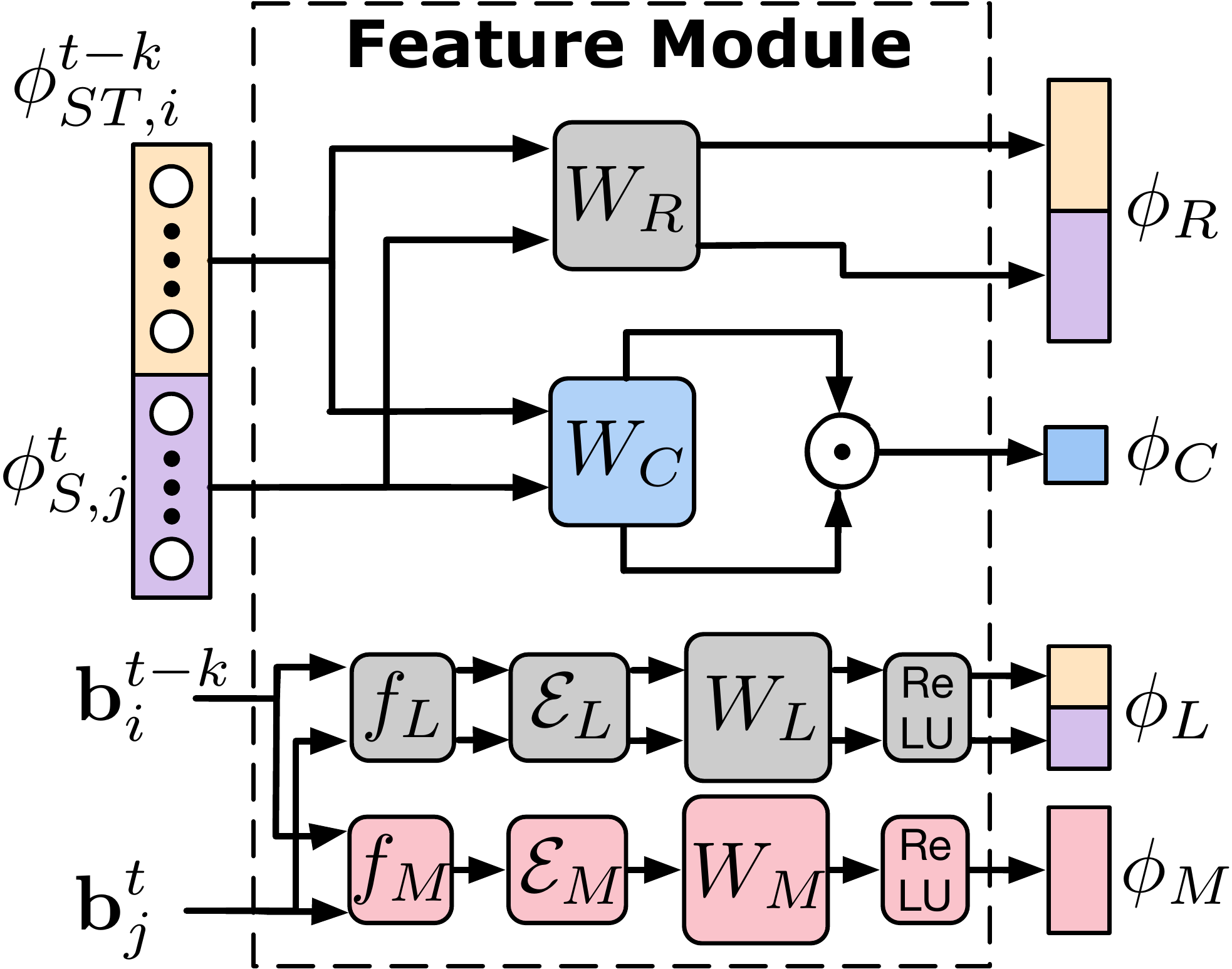}
    \caption{ The design of tracklet-object feature representation. $\phi_R$, $\phi_C$, $\phi_L$ and $\phi_M$ indicate the relation feature, cosine similarity, location feature and motion feature, respectively. All the features will be concatenated to produce the final similarity scores through a two-layer network with a sigmoid function.}
    \label{fig:feature-module}
\end{figure}

The cosine value is taken as an additional 1-d feature and fed to the following network for final similarity computation. The generation of hybrid relation features are summarized in Figure~\ref{fig:feature-module} (top).

In general, cosine value could take effect only in the scenarios where two input features are compatible in representation. At a first glance, it is not applicable to our ``incompatible'' features. Nevertheless, the features of tracklets and objects are actually compatible in some sensible way. The temporal relation in Eqn.~(\ref{eqn:temporal_relation}) is basically a weighted average over features from multiple frames. There is no projection between the object feature and tracklet feature. Hence, they still locate at a close space and are suitable to be modeled by cosine value.

In the experiments, the hybrid representation of pair relation features achieves superior accuracy than the methods using each of the formulations alone.

\vspace{-1.0em}
\subsubsection{Location Features}
Location/motion feature is another widely used cues in building the similarity score. We take the location/motion features from the last frame of a tracklet to represent the entire one, because the location/motion model in distant frames may drift a lot from the current frame.

The location features can be conveniently incorporated in our pipeline. The bare location features are firstly embedded and projected to the higher-dimensional space and then concatenated with the relation features to producing the final similarity score.  

We embed and project of bare location features follow~\cite{vaswani2017attention, hu2017relation} as
\begin{small}
\begin{equation}
\label{eqn:geo_embedding}
    {\phi _{*}} = W_{*} \cdot {\mathcal {E}_{*}\left(f_{*} ({\bf b}^{t-k}_i, {\bf b}^{t}_j) \right)},
\end{equation}
\end{small}
where $* \in \{L, M\}$ denotes the studied two types of location features, location and motion.
The first one is the normalized absolute location of bounding box:
\begin{small}
\begin{equation}
\label{eqn:absolution_location}
	f_L'\left( {{\textbf{b}}_j^t} \right) = \left( {\frac{{{x_j^t}}}{{I_{w}^t}},\frac{{{y_j^t}}}{{I_{h}^t}},\frac{{{w_j^t}}}{{I_{w}^t}},\frac{{{h_j^t}}}{{I_{h}^t}}} \right),
\end{equation}
\end{small}
where $I_{w}^t$ and $I_{h}^t$ are the width and height of frame $t$. $f_L$ in Eqn \ref{eqn:geo_embedding} is defined as $f_L\left( {\textbf{b}}_i^{t-k}, {{\textbf{b}}_j^t} \right) = \left[f_L'({\textbf{b}}_i^{t-k});f_L'({\textbf{b}}_j^t)\right]$.

The above location feature relates to the low-velocity assumption of objects, which has been proved work surprisingly well in the recent work~\cite{bochinski2017high}. Rather than using a hard constraint that the same objects on consecutive frames should have overlap, we incorporate the constraint softly into the feature representation, and the location patterns are learned from the data. 
The other location feature depict the motion information of an object in consecutive frames:
\begin{equation}
\label{eqn:motion_vector}\footnotesize
	f_M\left( {{\textbf{b}}_i^{t-k}, {\textbf{b}}_j^{t}} \right) = \log \left( {\frac{{{|x_i^{t-k} - x_j^{t}|}}}{{kw_i^{t-k}}},\frac{{{|y_i^{t-k} - y_j^{t}|}}}{{kh_i^{t-k}}},\frac{{{w_j^{t}}}}{{kw_i^{t-k}}},\frac{{{h_j^{t}}}}{{kh_i^{t-k}}}} \right).
\end{equation}
This location (motion) feature relates to the constant velocity assumption of objects, which is proved as a effective information for a robust similarity score.

\section{Experiments}

\subsection{Datasets and Evaluation Metrics}

We utilize three MOT Benchmarks \cite{MOTChallenge2015, MOT16} for evaluation. The benchmarks are challenging due to the large variety in frame rates, resolution, viewpoint, weather, camera motion and etc. These benchmarks are widely used in the field of multi-object tracking to evaluate different trackers.

\textbf{2D MOT2015  }consists of 11 training sequences and 11 testing sequences \cite{MOTChallenge2015}. Following~\cite{sadeghian2017tracking}, we split the training sequences into two subset of 4 training and 6 validation sequences for ablation study.

\textbf{MOT16  }consists of 7 training sequences and 7 testing sequences. The scenes are mostly crowd pedestrians and are regarded as more challenging.

\textbf{MOT17  }use the same videos as the \emph{MOT16} datasets but with better annotation and public detectors. All sequences are provided with three sets of detection results (DPM \cite{ross2010dpm}, Faster-RCNN \cite{ren2015faster} and SDP \cite{yang2016sdp}) for more comprehensive comparison of different multi-object trackers.

For a fair comparison, we use the public detection result provided with datasets as the input of our approach.

\vspace{-1em}
\paragraph{Evaluation Metric}
We adopt the standard metrics of MOT Benchmarks \cite{MOTChallenge2015, MOT16} for evaluation, including
Multiple Object Tracking Accuracy (MOTA) \cite{clearMOTmetric},
Multiple Object Tracking Precision (MOTP) \cite{clearMOTmetric},
ID F1 Score (IDF1, the ratio of correctly identified detections over the average number of ground-truth and computed detections) \cite{idf1},
ID Precision (IDP, the fraction of detected identities correctly identified), \cite{idf1},
ID Recall (IDR, the fraction of ground truth identities correctly identified), \cite{idf1},
Mostly tracked targets (MT, the ratio of ground-truth trajectories covered by an output trajectory for at least 80\% of ground truth length),
Mostly lost targets (ML, the ratio of ground-truth trajectories covered by an output trajectory for at most 20\% of ground truth length),
the number of False Positives (FP),
the number of False Negatives (FN),
the number of Identity Switch (IDS) \cite{idsw},
the number of Fragment Error (Frag).
The latest Average Ranking (AR) on the MOT benchmark website is also reported, which is computed by taking the average of benchmark ranking of all metrics above.

\subsection{Implementation Details}

\paragraph{Network Architecture}
We use ResNet-50~\cite{he2015resnet} as our backbone network. We first train it on ImageNet Image Classification task \cite{ILSVRC15} and then finetune the model on the MOT training datasets.

Given the bounding boxes of public detection, we crop and resize them to the resolution of $128 \times 64$. The cropped images are fed into the backbone network, producing a feature map with the resolution of $4\times 2$. a new 256-d $1\times 1$ convolution is applied on this feature map to reduce the channel dimension. A fully connected layer with dimension 1024 is applied right after the new $1\times 1$ conv layer, which is used as the representing appearance feature $\phi_i$ (see Section~\ref{sec:relation}).

In the spatial-temporal relation module, 
we mainly follow~\cite{hu2017relation, vaswani2017attention} for the hyper-parameters of spatial relation reasoning. For temporal relation, the object features from the latest 9 frames are aggregated.

After the relation module, pairing relation features and location features are extracted. The linear layers $W_R$, $W_c$ are of dimension 32 and 128, respectively. The function $\mathcal{E}_L$ embeds the 4-d bare location features to 64-d, followed by a linear layer $W_L$ to project the feature to 16-d. All of the relation features and location features are concatenated, forming a 65-d feature and fed to a two-layer network with a sigmoid function.

\vspace{-1em}
\paragraph{Training}
During training, all detection bounding boxes in input frames are cropped and fed into the network.
On average, each mini-batch contains 45 cropped images. 
A total of $437$k, $425$k and $1,275$k iterations are performed for 2DMOT2015, MOT16, MOT17 respectively. 
The learning rate is initialized as $10^{-3}$ and then decayed to $10^{-4}$ in the last $\frac{1}{3}$ training. Online hard example mining(OHEM) was adopt to address the heavy imbalance of positive/negative issue.

\vspace{-1em}
\paragraph{Inference}
In inference, the similarities between tracklets and objects on the current frame are computed according to Section 3.2. The association is then achieved by solving the bipartite graph as in Figure \ref{fig:pipeline}. 

Following the common practice for online tracking approaches~\cite{xiang2015learning, zhu2018online, chu2017online, sadeghian2017tracking}, we consider the too short tracklets as false alarms. Specifically, for a sequence with the frame rate of $F$, we remove the short tracklets if it is matched less than $0.3F$ times in the past $F$ frames after the initial match. Besides, we only keep the sequences that show up in the nearest $1.25F$ frames for enabling efficient inference.

\begin{table}[]
    \footnotesize
    \small
    \addtolength{\tabcolsep}{-4.3pt}
\begin{tabular}{lcccccccc}
\Xhline{1.0pt}
Feature    & MOTA & MOTP & IDF  & MT($\%$) & ML($\%$)  & FP   & FN    & IDS \\
\hline
A$_u$ & 19.8 & 72.3 & 26.2 & 4.7 & 53.4 & 1,800 & 14,309 & 2,177 \\
A$_c$ & 25.2 & 72.5 & 32.5 & 8.1 & 55.1 & 2,474 & 14,368 & 726 \\
A & 29.8 & 72.2 & 38.6 & 9.8 & 49.6 & 2,734 & 12,956 & 515 \\
\hline
A+L$_u$ & 31.7 & 72.7 & 40.8 & 8.5 & 54.2 & 1,477 & 13,946 & 355 \\
A+L$_m$ & 31.0 & 72.5 & 44.1 & 9.0 & 54.3 & 1,971 & 13,801 & 167 \\
A+L & 32.3 & 72.3 & 47.1 & 8.1 & 52.6 & 2,004 & 13,496 & 129 \\
\Xhline{1.0pt}
\end{tabular}
    \caption{Ablation study of various design of feature representation.}
\label{table:ablation-feature-module}
	\vspace{-2pt}
\end{table}

\begin{table}[]
    \small
    \centering
    \addtolength{\tabcolsep}{-5pt}
\begin{tabular}{lcccccccc}
\Xhline{1.0pt}
Module    & MOTA & MOTP & IDF  & MT($\%$) & ML($\%$)  & FP   & FN    & IDS \\
\hline
A+L & 32.3 & 72.3 & 47.1 & 8.1 & 52.6 & 2,004 & 13,496 & 129 \\
A+L+S & 34.8 & 72.4 & 46.5 & 9.0 & 53.0 & 947 & 13,966 & 151 \\
A+L+S+T & 36.2 & 72.2 & 46.6 & 9.0 & 51.7 & 1799 & 13,079 & 94 \\
\hline
A+L+S+Avg & 33.1  & 72.2 & 37.1 & 6.4 & 54.7 & 888 & 14,386 & 176 \\
A+L+S+Max & 33.9 & 72.4 & 43.4 & 8.5 & 54.7 & 848 & 14,268 & 140 \\
\Xhline{1.0pt}
\end{tabular}
    \caption{Ablation study of the spatial temporal relation network.}
\label{table:ablation-spatial-temporal-module}
	\vspace{-2pt}
\end{table}

\subsection{Ablation Study} \label{sec:ablation-exp}
We follow~\cite{sadeghian2017tracking} to split the 11 training sequences into train/val sets for ablation study.

\begin{table*}[tb]
    \centering
    \small
    \begin{minipage}{1.\textwidth}
    \centering
    \addtolength{\tabcolsep}{-3.5pt}
    \caption{Tracking Performance on 2DMOT2015 benchmark dataset.}
    \label{table:mot15}
\begin{tabular}{c|c|cccccccccccc}
        \Xhline{1.0pt}
Mode    & Method            & MOTA$\uparrow$ & MOTP$\uparrow$ & IDF$\uparrow$  & IDP$\uparrow$  & IDR$\uparrow$  & MT(\%)$\uparrow$      & ML(\%)$\downarrow$      & FP$\downarrow$     & FN$\downarrow$      & IDS$\downarrow$   & Frag$\downarrow$   & AR$\downarrow$   \\
\hline
\multirow{4}{30pt}{\centering Offline}
        & MHT\_DAM \cite{Kim15tracking} & 32.4 & 71.8 & \bluebold{45.3} & 58.9 & 36.8 & 16.0 & 43.8 & 9,064 & 32,060 & \bluebold{435} & 826 & 21.7 \\
        & NOMT \cite{Choi15aggregate} & 33.7 & 71.9 & 44.6 & \bluebold{59.6} & 35.6 & 12.2 & 44.0 & \bluebold{7,762} & 32,547 & 442 & \bluebold{823} & \bluebold{18.7} \\
        & QuadMOT \cite{son2017multi} & 33.8 & \bluebold{73.4} & 40.4 & 53.5 & 32.5 & 12.9 & \bluebold{36.9} & 7,898 & 32,061 & 703 & 1,430 & 23.5 \\
        & JointMC \cite{Keuper16tracking} & \bluebold{35.6} & 71.9 & 45.1 & 54.4 & \bluebold{38.5} & \bluebold{23.2} & 39.3 & 10,580 & \bluebold{28,508} & 457 & 969 & 19.3 \\
\hline
\multirow{5}{30pt}{\centering Online}
        & SCEA \cite{yoon2016scea} & 29.1 & 71.1 & 37.2 & 55.9 & 27.8 & 8.9 & 47.3 & 60,60 & 36,912 & \redbold{604} & \redbold{1,182} & 30.4 \\
        & MDP \cite{xiang2015learning} & 30.3 & 71.3 & 44.7 & 57.8 & 36.4 & 13.0 & 38.4 & 9,717 & 32,422 & 680 & 1,500 & 25.9 \\
        & CDA\_DDAL \cite{Bae18robust} & 32.8 & 70.7 & 38.8 & 58.2 & 29.1 & 9.7 & 42.2 & 4,983 & 35,690 & 614 & 1,583 & 24.2 \\
        & AMIR15 \cite{sadeghian2017tracking} & 37.6 & 71.7 & {46.0} & 58.4 & \redbold{38.0} & \redbold{15.8} & \redbold{25.8} & 7,933 & \redbold{29,397} & 1,026 & 2,024 & 19.6 \\
        & ours & \redbold{38.1} & \redbold{72.1} & \redbold{46.6} & \redbold{63.9} & 36.7 & 11.5 & 33.4 & \redbold{5,451} & 31,571 & 1,033 & 2,665 & \redbold{16.1} \\
\Xhline{1.0pt}
\end{tabular}
\end{minipage}

\begin{minipage}{1.\textwidth}
\centering
\addtolength{\tabcolsep}{-3.pt}
    \caption{Tracking Performance on MOT16 benchmark dataset.}
    \label{table:mot16}
\begin{tabular}{c|c|cccccccccccc}
        \Xhline{1.0pt}
Mode    & Method            & MOTA$\uparrow$ & MOTP$\uparrow$ & IDF$\uparrow$  & IDP$\uparrow$  & IDR$\uparrow$  & MT(\%)$\uparrow$      & ML(\%)$\downarrow$      & FP$\downarrow$     & FN$\downarrow$      & IDS$\downarrow$   & Frag$\downarrow$   & AR$\downarrow$   \\
\hline
\multirow{6}{30pt}{\centering Offline}
        & NOMT \cite{Choi15aggregate}     & 46.4 & 76.6 & \bluebold{53.3} & 73.2 & \bluebold{41.9} & 18.3 & 41.4 & 9,753  & 87,565  & \bluebold{359}   & \bluebold{504}   & 18.6 \\
        & MCjoint \cite{Keuper16tracking} & 47.1 & 76.3 & 52.3 & \bluebold{73.9} & 40.4 & \bluebold{20.4} & 46.9 & 6,703  & 89,368  & 370   & 598   & 19.8 \\
        & NLLMPa \cite{Levinkov17nllmpa}  & 47.6 & 78.5 & 47.3 & 67.2 & 36.5 & 17.0 & 40.4 & 5,844  & 89,093  & 629   & 768   & 18.8 \\
        & FWT \cite{Henschel17headbody}   & 47.8 & 75.5 & 44.3 & 60.3 & 35   & 19.1 & \bluebold{38.2} & 8,886  & \bluebold{85,487}  & 852   & 1,534 & 24.8 \\
        & GCRA \cite{Ma18GCRA}  & 48.2 & 77.5 & 48.6 & 69.1 & 37.4 & 12.9 & 41.1 & \bluebold{5,104}  & 88,586  & 821   & 1,117 & 21.9 \\
        & LMP \cite{tang2017multiple}     & \bluebold{48.8} & \bluebold{79.0}   & 51.3 & 71.1 & 40.1 & 18.2 & 40.1 & 6,654  & 86,245  & 481   & 595   & \bluebold{17.8}\\
\hline
\multirow{6}{30pt}{\centering Online}
	    & oICF \cite{Kieritz2016avss}    & 43.2 & 74.3 & 49.3 & 73.3 & 37.2 & 11.3 & 48.5 & 6,651  & 96,515  & \redbold{381}   & \redbold{1,404} & 31.8 \\
        & STAM \cite{chu2017online}    & 46.0   & 74.9 & 50   & 71.5 & 38.5 & 14.6 & 43.6 & 6,895  & 91,117  & 473   & 1,422 & 29.3 \\
        & DMAN \cite{zhu2018online}     & 46.1 & 73.8 & \redbold{54.8} & \redbold{77.2} & 42.5 & \redbold{17.4} & 42.7 & 7,909  & 89,874  & 532   & 1,616 & 23.4 \\
        & AMIR \cite{sadeghian2017tracking}    & 47.2 & \redbold{75.8} & 46.3 & 68.9 & 34.8 & 14.0 & 41.6 & \redbold{2,681}  & 92,856  & 774   & 1,675 & 22.9 \\
        & MOTDT \cite{Chen18reid} & 47.6 & 74.8 & 50.9 & 69.2 & 40.3 & 15.2 & 38.3 & 9,253  & 85,431  & 792   & 1,858 & 23.5 \\
        & ours     & \redbold{48.5} & 73.7 & 53.9 & 72.8 & \redbold{42.8} & 17.0 & \redbold{34.9} & 9,038 & \redbold{84,178} & 747 & 2,919 & \redbold{15.4} \\
        \Xhline{1.0pt}
\end{tabular}
\end{minipage}

\begin{minipage}{1.\textwidth}
\centering
\addtolength{\tabcolsep}{-4.08pt}
    \caption{Tracking Performance on MOT17 benchmark dataset.}
    \label{table:mot17}
\begin{tabular}{c|c|cccccccccccc}
\Xhline{1.0pt}
Mode    & Method            & MOTA$\uparrow$ & MOTP$\uparrow$ & IDF$\uparrow$  & IDP$\uparrow$  & IDR$\uparrow$  & MT(\%)$\uparrow$      & ML(\%)$\downarrow$      & FP$\downarrow$     & FN$\downarrow$      & IDS$\downarrow$   & Frag$\downarrow$   & AR$\downarrow$   \\
\hline
\multirow{6}{30pt}{\centering Offline}
 & IOU \cite{Bochinski17highspeed}        & 45.5 & 76.9 & 39.4 & 56.4 & 30.3 & 15.7 & 40.5 & \bluebold{19,993} & 281,643 & 5,988 & 7,404  & 36.5 \\
  & MHT\_DLSTM \cite{kim2018multi}     & 47.5 & \bluebold{77.5} & 51.9 & 71.4 & 40.8 & 18.2 & 41.7 & 25,981 & 268,042 & 2,069 & 3,124 & 28.8 \\
  & EDMT \cite{Chen17mht}       & 50.0   & 77.3 & 51.3 & 67   & 41.5 & \bluebold{21.6} & 36.3 & 32,279 & \bluebold{247,297} & 2,264 & 3,260  & 24.0  \\
  & MHT\_DAM \cite{Kim15tracking}   & 50.7 & \bluebold{77.5} & 47.2 & 63.4 & 37.6 & 20.8 & 36.9 & 22,875 & 252,889 & 2,314 & \bluebold{2,865}  & 25.4 \\
  & jCC \cite{Keuper18jcc}   & 51.2 & 75.9 & \bluebold{54.5} & \bluebold{72.2} & \bluebold{43.8} & 20.9 & 37   & 25,937 & 247,822 & \bluebold{1,802} & 2,984 & 、\bluebold{20.3} \\
  & FWT \cite{Henschel17headbody}   & \bluebold{51.3} & 77   & 47.6 & 63.2 & 38.1 & 21.4 & \bluebold{35.2} & 24,101 & 247,921 & 2,648 & 4,279 & 24.2 \\
\hline
\multirow{6}{30pt}{\centering Online}
  & PHD\_GSDL \cite{Fu18PHDGSDL}     & 48.0 & \redbold{77.2} & 49.6 & 68.4 & 39 & 17.1 & \redbold{35.6} & 23,199 & 265,954 & 3,998 & 8,886 & 32.5 \\
  & AM\_ADM \cite{Lee18AM}     & 48.1 & 76.7 & 52.1 & 71.4 & 41 & 13.4 & 39.7 & 25,061 & 265,495 & 2,214 & 5,027 & 27.3 \\
  & DMAN \cite{zhu2018online}   & 48.2 & 75.9 & 55.7 & \redbold{75.9} & 44   & 19.3 & 38.3 & 26,218 & 263,608 & 2,194 & 5,378  & 26.6 \\
  & HAM\_SADF \cite{Yoon18historical}     & 48.3 & \redbold{77.2} & 51.1 & 71.2 & 39.9 & 17.1 & 41.7 & \redbold{20,967} & 269,038 & \redbold{1,871} & \redbold{3,020} & 25.2 \\
  & MOTDT \cite{Chen18reid} & \redbold{50.9} & 76.6 & 52.7 & 70.4 & 42.1 & 17.5 & 35.7 & 24,069 & 250,768 & 2,474 & 5,317 & 23.1 \\
  & ours     & \redbold{50.9} & 75.6 & \redbold{56.5} & 74.5 & \redbold{45.5} & \redbold{20.1} & 37.0 & 27,532 & \redbold{246,924} & 2,593 & 9,622 & \redbold{18.2}  \\
        \Xhline{1.0pt}
\end{tabular}
\end{minipage}
	\vspace{-10pt}
\end{table*}

\vspace{-1em}
\paragraph{Design of Feature Representation} We first examine the effects of various design of feature representation in Table~\ref{table:ablation-feature-module}. All the experiments are based on the original appearance features without spatial-temporal reasoning.

The first three rows compare the effects of different appearance features \emph{without the relation modules}. By only using unary appearance representation, it achieves 19.8 in terms of MOTA. By using cosine value alone, it gets 25.2 in MOTA. By using the hybrid features of both unary appearance and cosine value, the accuracy is significantly higher, reaching 29.8 in MOTA.

The last three rows compare the effects of different location features. By only utilizing the unary location features in Eqn.~(\ref{eqn:absolution_location}), 1.9 MOTA improvements is observed. By utilizing the motion features in Eqn.~(\ref{eqn:motion_vector}), 1.2 improvements in MOTA is observed. By combining both of them, we achieve 2.5 MOTA boosts. Also note that with the location features, the ID switch is significantly reduced, from 515 to 129.

\vspace{-1em}
\paragraph{The effects of Spatial-Temporal Relation Module}
Table~\ref{table:ablation-spatial-temporal-module} examines the effects of spatial-temporal relation module in improving the tracking accuracy. With relation reasoning along the spatial domain, the tracking accuracy improves by 2.5 in terms of MOTA. Significant reduction in FP is observed, indicating the topology encoded by spatial relation reasoning could help the association method to more accurately identify wrong associations. Further performing temporal relation reasoning, an additional 1.4 MOTA improvement is achieved. Note that our temporal relation reasoning is essentially a weighted average over all frame features. Hence we also compare it to some straight-forward aggregation methods, such as average summation and max-pooling along the frame dimension. These methods perform significantly worse than ours, proving the effectiveness of our temporal relation reasoning method.

\subsection{Results on the MOT Benchmarks}

We report the tracking accuracy on all of the three MOT benchmarks in Table~\ref{table:mot15}, \ref{table:mot16} and \ref{table:mot17}. We used the public detections for a fair comparison. Our method achieves the state-of-the-art tracking accuracy under online settings on all of the three benchmarks considering the major metrics of MOTA and AR (average rank).

\section{Conclusion}

This paper studies the object association problem for multi-object tracking (MOT). To build a robust similarity measure, we combine various cues, including appearance, location and topology cues through utilizing relation networks in spatial domains and further extending the relation networks to the temporal domain for aggregating information across time. The resulting approach is dubbed as spatial-temporal relation networks (STRN), which runs feed-forward and in end-to-end. It achieves the state-of-the-art accuracy over all online methods on all of the MOT15$\sim$17 benchmarks using public detection.

{\small
\bibliographystyle{ieee}
\bibliography{MOT}
}

\end{document}